\documentclass[runningheads]{llncs}
\usepackage[english]{babel}
\usepackage[utf8]{inputenc}

\usepackage{amsfonts}
\usepackage{amsmath}
\usepackage{amssymb}
\usepackage{mathtools}
\usepackage{bbm}

\usepackage{graphicx}
\usepackage{hyperref}
\usepackage{pgfplots}

\usepackage{tikz}
\usepackage{physics}

\usepackage{calligra}
\usepackage{csquotes}

\usepackage{tensor}
\usepackage[thicklines]{cancel}

\usepackage{color}
\usepackage{xcolor}
\usepackage{tcolorbox}

\usepackage{pstricks}

\usepackage{algorithm}
\usepackage{algorithmic}
\usepackage{listings}
\usepackage{mathtools}

\usepackage{booktabs}
\usepackage{multirow}

\pgfplotsset{compat=1.17}
\definecolor{blue2}{HTML}{045FB4}
\definecolor{green2}{HTML}{46C235}
\definecolor{red2}{HTML}{EE4848}
\definecolor{violet2}{HTML}{A647E5}
\definecolor{orange2}{HTML}{FF7425}
\definecolor{darkred}{HTML}{5C2020}
\definecolor{gray}{HTML}{303030}
\definecolor{yellow}{HTML}{f0be52}
\definecolor{lightdarkgold}{HTML}{EEBC1D}

\newcommand{\mtx}[2]{\ensuremath{\textbf{#1}_\textbf{#2}}}
\newcommand{\vtr}[2]{\ensuremath{\textbf{#1}_\textbf{#2}}}
\newcommand{\w}[1]{\ensuremath{\textbf{W}_\textbf{#1}}}

\newcommand{\x}[1]{\ensuremath{\textbf{x}_\textbf{#1}}}

\newcommand{\bias}[1]{\ensuremath{\textbf{b}_\textbf{#1}}}

\newcommand{\y}[1]{\ensuremath{y_{#1}}}

\newcommand{\quant}[2]{\ensuremath{Q\prnt{#1, #2}}}
\newcommand{\ce}[1]{\ensuremath{\mc{H}_{\mathrm{CE}}^{#1}}}

\newcommand{\ceg}{CEG4N}

\newtheorem{assumption}{Assumption}

\DeclareMathOperator*{\argmax}{\arg\max}
\DeclareMathOperator*{\argmin}{\arg\min}

\DeclareMathAlphabet{\mathcalligra}{T1}{calligra}{m}{n}
\DeclareFontShape{T1}{calligra}{m}{n}{<->s*[2.2]callig15}{}

\newcommand{\ie}{\emph{i.e.}} 
\newcommand{\eg}{\emph{e.g.}} 
\newcommand{\nn}{NN}
\newcommand{\ns}{NNs}

\newcommand{\te}{Top-1-Equivalence~}

\newcommand{\mc}[1]{\ensuremath{\mathcal{#1}}}

\newcommand{\round}[1]{\ensuremath{\left\lfloor#1\right\rceil}}
\newcommand{\act}[1]{\ensuremath{\sigma\left(#1\right)}}

\newcommand{\Eq}[1]{Eq.~\ref{#1}}

\newcommand{\prnt}[1]{\ensuremath{\left(#1\right)}} 
\newcommand{\sbkt}[1]{\ensuremath{\left[#1\right]}} 
\newcommand{\cbkt}[1]{\ensuremath{\left\{#1\right\}}}  

\begin{document}
\title{\ceg: Counter-Example Guided Neural Network Quantization Refinement}
%
%
\author{João Batista P. Matos Jr.\inst{1}
\and Iury Bessa\inst{1}
\and Edoardo Manino\inst{2}
\and Xidan Song\inst{2}
\and Lucas C. Cordeiro\inst{2,1}
}
\authorrunning{Matos Jr. et al.}
%
\institute{Federal University of Amazonas, Manaus-AM, Brazil \\
\email{jbpmj@icomp.ufam.edu.br} and
\email{iurybessa@ufam.edu.br}\\
\and Univeristy of Manchester, Machester, United Kingdom
\email{\{lucas.cordeiro,eduardo.manino,xidan.song\}@manchester.ac.uk}
}

\maketitle              
%


\begin{abstract}

Neural networks are essential components of learning-based software systems. However, their high compute, memory, and power requirements make using them in low resources domains challenging. For this reason, neural networks are often quantized before deployment. Existing quantization techniques tend to degrade the network accuracy. We propose Counter-Example Guided Neural Network Quantization Refinement (\ceg). This technique combines search-based quantization and equivalence verification: the former minimizes the computational requirements, while the latter guarantees that the network's output does not change after quantization. We evaluate \ceg~on a diverse set of benchmarks, including large and small networks. Our technique successfully quantizes the networks in our evaluation while producing models with up to 72\% better accuracy than state-of-the-art techniques. 

\end{abstract}

\section{Introduction}

Neural networks (\ns) are becoming essential in many applications such as autonomous driving~\cite{Bojarski2016EndTE}, security, medicine, and business~\cite{ABIODUN2018e00938}. However, current state-of-the-art \ns~often require substantial compute, memory, and power resources, limiting their applicability~\cite{Cheng2017ASO}.

In this respect, quantization techniques help reduce the network size and its computational requirements~\cite{Cheng2017ASO,Kirchhoffer2021,Gholami2022ASO}. Here, we focus on \textit{quantization} techniques, which aim at reducing the number of bits required to represent the neural network weights~\cite{Gholami2022ASO}. A desirable quantization technique produces the smallest neural network possible from the quantization perspective. However, at the same time, quantization affects the functional behavior of the resulting neural network by making them more prone to erratic behavior due to loss of accuracy~\cite{Hooker2019SelectiveBD}. For this reason, existing techniques monitor the degradation in the accuracy of the quantized model with statistical measures defined on the training set~\cite{Gholami2022ASO}.

However, statistical accuracy measures do not capture the network's vulnerability to malicious attacks. Indeed, there may exist some specific inputs for which the network performance degrades significantly~\cite{Huang2018SafetyAT,Liu2021AlgorithmsFV,Albarghouthi2021IntroductionTN}. For this reason, we reformulate the goal of guaranteeing the accuracy of a quantized model under the notion of \textit{equivalence}~\cite{Farabet2011LargeScaleFC,Han2015LearningBW,NIPS2015_10a5ab2d,Hubara2017QuantizedNN}. This formal property requires that two neural network models both produce the same output for every input, thus ensuring that the two networks are functionally equivalent~\cite{Narodytska2018VerifyingPO,Teuber2021Equiv}.

We are the first to explore the combination of quantization techniques and equivalence checking in the present work. Doing so guarantees that the quantized model is functionally equivalent to the original one. More specifically, our main scientific contributions are the following:
\begin{itemize}
    \item We model the equivalence quantization problem as an iterative optimization-verification cycle.
    \item We propose \ceg, a counter-example guided neural network quantization technique that provides formal guarantees of \nn~equivalence.
    \item We evaluate \ceg~on both large (ACAS Xu~\cite{Julian2016} and MNIST~\cite{LeCun2005TheMD}) and small (Iris~\cite{IrisDataset} and Seeds~\cite{Charytanowicz}) benchmarks.
    \item We demonstrate that \ceg~can successfully quantize neural networks and produce models with similar or better accuracy than a baseline state-of-the-art quantization technique (up to 72\% better accuracy).
\end{itemize}


\section{Preliminaries}
\label{sec:preliminaries}

\subsection{Neural Network}
\label{sec:neural-network}

\ns~are non-linear mapping functions  $f: \mc{I} \subset \mathbb{R}^n \rightarrow \mc{O} \subset \mathbb{R}^m$ consisting of a set of $L$ linked layers, organized as a \emph{direct graph}. Each layer $l$ is connected with the directly preceding layer $l-1$, \ie, the output of the layer $l-1$ is the input of the layer $l$. Exceptions are the first and last layers. The first layer is just a placeholder for the input for the \nn~while the last layer holds the \nn~function mapping $f$. 
A layer $l$ is composed by a matrix of weights $\mtx{W}{l}\in\mathbb{R}^{n\times m}$ and a bias vector $\vtr{b}{l}\in\mathbb{R}^m$. 

The output of a layer is computed by performing the combination of an affine transformation, followed by the non-linear transformation on its input $\vtr{x}{l} \in \mathbb{R}^n$(see Eq.~\eqref{eq:layer}). Formally, we can describe the function $y_{l}:\mathbb{R}^n \rightarrow \mathbb{R}^m$ that computes the output of a layer $l$ as follows: 



\begin{equation}
\label{eq:layer}
   y_{l}(\x{l}) = \mtx{W}{l}\cdot\vtr{x}{l} + \vtr{b}{l}
\end{equation}
and the function that computes the activated output of a layer $l$ as follows:
\begin{equation}
\label{eq:activation}
   y_{l}^\sigma(\x{l}) = \sigma(y_{l}(\x{l}) )
\end{equation}
where $\sigma:\mathbb{R}^m\rightarrow \mathbb{R}^m$ is the \emph{activation function}. In other words, the output $l$ is the result of the activation function $\sigma$ applied to the dot product between weight and input, plus the bias. The most popular activation functions are: namely, ReLU, sigmoid (Sigm), and the re-scaled version of the latter known as hyperbolic tangent(TanH)~\cite{Eleftheriadis2022OnNN}. We focus on the \emph{rectified linear unit} activation function $ReLU=\max\cbkt{0, \vtr{y}{l}}$. 

Considering the above, let us denote the input of a \nn~with $L$ layers as $\x~\in\mathbb{I}$~, and $f(x)\in\mc{O}$ as the output; thus, we have that:
 \begin{equation}
 \label{eq:network}
   f(\x{}) = \act{{y_L}(\act{{y_{L-1}}(...(\act{{y_1(\x{})}})))})}\\
 \end{equation}

\subsection{Quantization}
\label{sec:quantization}

Quantization is the process of constraining high precision values (\eg, single-precision floating-point values) to a finite range of lower precision values (\eg, a discrete set such as the integers)~\cite{AbateBCCDKK17,Gholami2022ASO}. The quantization quality is usually determined by a scalar $n$ (the available number of bits) that defines the lower and upper bounds of the finite range. Let us define quantization as a mapping function $\mc{Q}_n: \mathbb{R}^{m\times p}\rightarrow\mathbb{I}^{m\times p}$, formulated as follow:
\begin{equation}
\label{eq:quantized-r-clip}
   \quant{n}{A} = clip\left(\round{\frac{A}{q(A, n)}}, -2^{n-1}, 2^{n-1} - 1\right)
\end{equation}

\noindent where $A\in \mathbb{R}^{m\times~p}$ denotes the continuous value-- notice that $A$ can be a single scalar, a vector, or a matrix; $n$ denotes the number of bits for the quantization, $q(A, n)$ denotes a function that calculates the scaling factor for $A$ in respect to a number of bits $n$, and \round{\cdot}~denotes rounding to the nearest integer. Defining the scaling factor (see \Eq{eq:scaling-factor}) is an important aspect of uniform quantization~\cite{Jacob2017Quantization,Krishnamoorthi2018Quantizing}. 

The scaling factor is essentially what divides a given range of real values $A$ into an arbitrary number of partitions. Thus, let us define a scaling factor function by $q_n(A)$, a number of bits (bit-width) to be used for the quantization by $n$, a clipping range by \sbkt{\alpha, \beta}, the scaling factor can be defined as follow:

\begin{equation}
\label{eq:scaling-factor}
q(A, n) = \frac{\beta-\alpha}{2^n -1}
\end{equation}

The min/max of the signal are often used to determining the clipping range values, \ie, $\alpha = \min{A}$ and $\beta = \max{A}$. But as we are using symmetric quantization, the clipping values are defined as $\alpha = \beta = \max([|\min{A}|, |\max{A}|])$. In practice, the quantization process can produce an integer value that lies outside the range of $[\alpha, \beta]$. To prevent that, the quantization process will have an additional clip step.



Eq.~\eqref{eq:dequantization} shows the corresponding de-quantization function, which computes back the original floating-point value. However, we should note that the de-quantization approximates the original floating-point value.
\begin{equation}
\label{eq:dequantization}
\hat{A} = q(A, 2)\quant{n}{A}
\end{equation}



\subsection{\nn~quantization}
\label{sec:nn-quantization}

In this section, we discuss how a convolutional or fully-connected \nn~layer can be quantized in the symmetric mode. Considering $l$ to be any given layer in a \nn, let us denote \x{l}, \w{l}, and \bias{l} as the original floating-point input vector, the original floating-point weight matrix, and the original floating-point bias vector, respectively, of the layer $l$. And applying the de-quantization function from Eq.~\eqref{eq:dequantization}, where, we assume that $A = \hat{A}$. Borrowing from notations used in Sections \ref{sec:neural-network} and \ref{sec:quantization}. We can formalize the quantization of a \nn~layer $l$ as follows:


\begin{equation}
\label{eq:quantized-layer}
\begin{split}
        \y{1}(\x{l}) & = {\w{l} \cdot\x{l} + \bias{l}}\\
        \quad &  \approx q(\w{l}, n_l)\quant{n_l}{\w{l}} \cdot \x{l} + q(\bias{l}, n_l)\quant{n_l}{\bias{l}}\\
\end{split}
\end{equation}


Notice that the bias does not need to be re-scaled to match the scale of the dot product. Since we consider maximum scaling factor between $q(\w{l}, n_l)$ and $q(\bias{l}, n_l))$, both the weight and the bias share the same scaling factor in Eq.~\eqref{eq:quantized-layer}. With that in mind, the formalization of a \nn $f$ in Eq.~\eqref{eq:network} can be reused to formalize a quantized \nn~as well.


\subsection{\nn~Equivalence}
\label{sec:nn-equivalence}

Let \mc{F} and \mc{T} be two arbitrary \ns, and let $\mc{I} \in \mathbb{R}^{n}$ be the common input space of the two \ns~and $\mc{O} \in \mathbb{R}^m$ be their common output space. Thus, \nn~equivalence verification is the problem of proving that \mc{F} and \mc{T}, or more specifically, their corresponding mathematical functions $f:\mc{I}\rightarrow\mc{O}$, $t:\mc{I}\rightarrow\mc{O}$ are equivalent. In essence, by proving the equivalence between two neural networks, one can prove that both \ns~ produce the same outputs for the same set of inputs. Currently, the literature reports the following definition of equivalence.

\begin{definition}[\te~\cite{BuningKS20Equivalence,Teuber2021Equiv}]
\label{def:top-1-equivalence}
Two \ns~$f$ and $t$ are Top-1-equivalent, if $\argmax~f(x) = \argmax~t(x)$, for all $x~\in~\mc{I}$.
\end{definition}

Let us formalise the notion of \emph{Top-1 Equivalence} in first-order logic. This is necessary for the comprehension of the equivalence verification explained in the following sections of the paper. But first, we formalize some essential assumptions for the correctness of the equivalence properties.

\begin{assumption}
Let $f(x)$ be the output of the \nn~\mc{F} in real arithmetic (without quantization). It is assumed that  $\argmax f(x) = y$ such that $x\in \mc{H}$.
\end{assumption}

\begin{assumption}
Let $f^q(x)$ be the output of the \nn~\mc{F} in a quantized form. There is set of numbers of bits \mc{N} such that $\argmax f(x) = \argmax f^q(x) = y$ for all $x\in \mc{H}$. 
\end{assumption}

Note that the quantization of the \nn~$f$ that results in the \nn~$f^q(x)$ depends on the number of bits $N$. Refer to Eq.~\eqref{eq:quantized-layer} to understand the relationship between \mc{N} and $f^q$.

An instance of a equivalence verification is given by a conjunction of constraints on the input $\psi_x(x)$, the output $\psi_y(y)$ and the \ns~$f$ and $f^q$. $\psi(f, f^q, x, y) = \psi_x(x)\rightarrow\psi_y(y)$. We denote $\psi_y(y)$ the equivalence constraint. Let $\bar{x} = x + \hat{x}$ such that $|x+\hat{x}|_{\infty} \leq \epsilon$, consider $\bar{x} \in \mc{H}$ and $y \in \mc{G}$. Taking from Definition~\ref{def:top-1-equivalence}, we have that:  

\begin{itemize}
    \item $\psi_x(x)$ is an equivalence property such that $\psi_x(x) \leftrightarrow \bar{x}\in \mc{H}$
    
    \item $\psi_y(y)$ is an equivalence property such that $\psi_y(y) \leftrightarrow  \argmax f^q(x) = y$
\end{itemize}


Note that, to prove the equivalence of $f$ and $f^q$, one may prove that the property $\psi(f, f^q,x, y)$ holds for any given $x$ and $y$. This approach may not be feasible. But proving that $\psi(f, f^q,x, y)$ does not hold for some $x$ and $y$ is a more tractable approach. If we do so, we can provide a counter-example.

\subsection{Verification of \nn~properties}
\label{subsec:prop_verify}

In this paper, we use the classic paradigm of SMT verification. In this paradigm, the property to check (e.g., equivalence) and the computational model (e.g., the neural networks) are encoded as a first-order logic formula, which is then checked for satisfiability. Moreover, to keep the problem decidable, SMT restricts the full expressive power of first-order logic to a decidable fragment. 

SMT formulas can capture the complex relationship between variables, holding real, integer values and other data types. If it is possible to assign values to such variables that a formula is evaluated as true, then the formula is said to be \emph{satisfiable}. On the other hand, if it's not possible to assign such values, the formula is said to be \emph{unsatisfiable}.

Given a \nn~\mc{F} and its mathematical function $f$, a set of safe input instances $\mc{H}\in\mathbb{R}^n$, and a safe domain $\mc{G}\subseteq\mc{O}^m$-- both defined as a set of constraints, safety verification is concerned with the question of whether there exist an instance $x\in\mc{H}$ such that $f(x)\notin\mc{G}$. An instance of a safety verification is given by a conjunction of constraints on the input $\psi_x(x)$, the output $\psi_y(y)$ and the \nn~$f$. $\psi(f, x, y) = \psi_x(x)\rightarrow\psi_y(y)$
is said to be satisfiable if there exists some $x\in\mathcal{H}$ such that $f(x)$ returns $y$ for the input $x$ and $\psi(f, x, y)$ does not hold. 



\section{Counter-Example Guided Neural Network Quantization Refinement (CEG4N)}

We define \emph{robust quantization (RQ)} to describe the problem of maximizing the quantization of a \nn~while keeping the equivalence between the original model and the quantized one (see Definition~\ref{def:robust-quantization}). Borrowing from the notations used in Section~\ref{sec:preliminaries}, we formally define RC as follows.

\begin{definition}[Robust Quantization\label{def:robust-quantization}]
Let $f$ be the reference \nn~ and $\mc{H}\in\mathbb{R}^n$ be a set of inputs instances. We define robust quantization as a process that performs the quantization of $f$ hence resulting in a quantized model $f^q$ such that $\argmax f(x) \iff \argmax f^q(x)~\forall~x\in\mc{H}$. 
\end{definition}

From the definition discussed in Section~\ref{sec:nn-equivalence}, we preserve the equivalence between the mathematical functions $f$ and $f^q$ associated with the \ns. In the RC, we shift the focus from the original \nn~to the quantized \nn, \ie, we assume that $f$ is safe (or robust) and use it as a reference to define the safety properties we expect for $f^q$. By checking the equivalence of $f$ and $f^q$, we can state that $f^q$ is robust, and therefore, we achieve a \textit{robust quantization}. In more details, consider a \nn~$f$ with $L$ layers. The quantization of $f$ assumes there is a set $\mc{N} = \cbkt{n_1, n_2, \cdots, n_L}$, where $n_l\in\mc{N}$ represents the number of bits that should be used to quantize the $l$-th layer in $f$. In our robust quantization problem, we obtain a sequence \mc{N} for which each $n \in \mc{N}$ is minimized (e.g., one could minimize the sum of all $n \in \mc{N}$) and the equality between $f$ and $f^q$ is satisfied.

\subsection{Robust quantization as a minimization problem}
\label{sec:robust-quantization}



We consider the robust quantization of a \nn~as an iterative minimization problem. Each iteration is composed of two complementary sub-problems. First, we need to minimize the quantization bit widths, that is, finding a candidate set \mc{N}. Second, we need to verify the equivalence property, that is, checking if a \nn~quantized with the bit widths in \mc{N} is equivalent to the original \nn. If the latter fails, we iteratively return to the minimization sub-problem with additional information. More specifically, we formalize the first optimization sub-problem as follows.

\textbf{Optimization sub-problem $o$: }
\begin{equation}
 \begin{split}
     \textbf{Objective: } \quad & \mathcal{N}^{o}=\argmin_{n_1^{o},\dots,n_L^{o}}~\sum_{l\in \mathbb{N}_{l\leq L}}{n_{l}}  \\
     \textbf{s.t: } \quad & \argmax f(x)=\argmax f^{q}(x),~\forall~ x\in\ce{o}\\
     \quad & n_l \geq \underline{N}~\forall~n_l \in\mc{N}^o \\
     \quad & n_l \leq \overline{N}~\forall~n_l \in\mc{N}^o 
 \end{split}
\end{equation}
\noindent where $f$ is the mathematical function associated with the \nn~\mc{F} and $f^q$ is the quantized mathematical function associated with the \nn~\mc{F}, \ce{o} is a set of counter-examples available at iteration $o$. Consider $\underline{N}$ and $\overline{N}$ as the minimum and the maximum bit width allowed to be used in the quantization; these parameters are constant. $\overline{N}$ ensures two things, it gives an upper bound to the quantization bit width, and provides a termination criteria, if a candidate $\mc{N}^o$ such that $n_l = \overline{N}$ for every $n_l \in \mc{N}^o$, the optimization is stopped because it reached our \textbf{Assumption 2}. In particular, our \textbf{Assumption 2} ensures the termination of \ceg, and it is build over the fact that there is a set of $\overline{N}$ for which the quantization introduces a minimal amount of error to \nn. In any case, if \ceg~proposes a quantization solution equal to the $\overline{N}$, this solution is verified as well, and in case the verification returns a counter-example, \ceg~finishes with failure. Finally, note that \ce{o} is an iterative parameter, meaning its value is updated at each iteration $o$. This is done based on the verification sub-problem (formalized below).   

\textbf{Verification sub-problem $o$:}

\noindent In the verification sub-problem $o$, we check whether the $\mathcal{N}^{o}$ generated by the optimization sub-problem $o$ satisfies the following equivalence property:
\begin{align*}
    \psi(f, f^q, x, y) = \psi_x(x)\rightarrow\psi_y(y)
\end{align*}
\noindent if $\psi_x(x)\rightarrow\psi_y(y)$ holds for the candidate $\mathcal{N}^{o}$, the optimization halts and $\mathcal{N}^{o}$ is declared as solution; otherwise, a new counter-example $x_\mathrm{CE}$ is generated. Iteration $o+1$ starts where iteration $o$ stopped. That is, the optimization sub-problem $o+1$ receives as parameter a set of \ce{o+1} such that $\ce{o+1} = \ce{o} \cup x_\mathrm{CE}$. 

\subsection{The \ceg~framework implementation}


We propose \ceg~framework, which is a counterexample-guided optimization approach to solve the robust quantization problem. In this approach, we consider combining two main modules to solve the two sub-problems presented in Section \ref{sec:robust-quantization}: the optimization of the bit widths for the quantization and the verification of the \nn~equivalence. The first module that solves the optimal bit width problem roughly takes in a \nn~and generates quantized~\nn~candidates. Then, the second module takes in the candidates and verifies their equivalence to the original model.

\begin{figure}[t]
\centering
\includegraphics[width=1.0\textwidth]{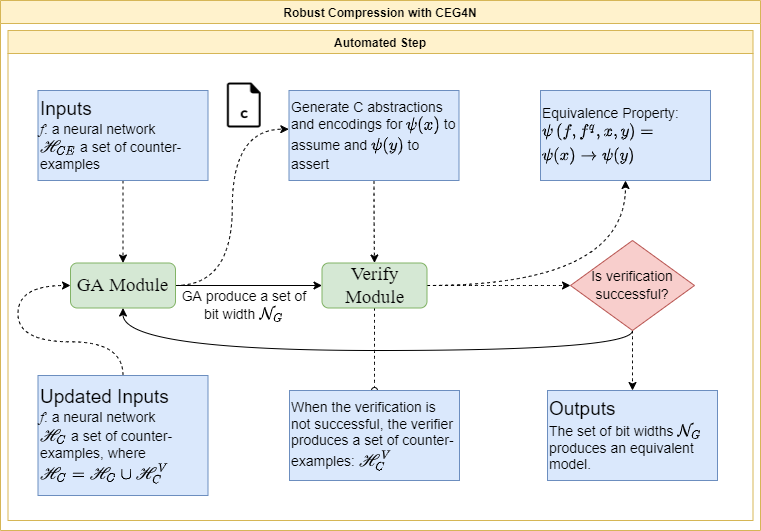}
\caption{\ceg~architecture overview, highlighting the relationship between the main modules, and their inputs and outputs. \label{fig:rc-cegio}}
\end{figure}

Figure~\ref{fig:rc-cegio} illustrates the overall architecture of the \ceg~framework. It also shows how each framework's module interacts with the other and in what sequence. The \textit{GA module} is an instance of a Genetic Algorithm. The GA module expects two main parameters, \nn~and a set of counter-examples~\ce. We can also specify a maximum number of generations the algorithm is allowed to run and lower and upper bounds to restrict the possible number of bits. Once the GA module produces a candidate, that is, a sequence of bit widths, for each layer of the neural network, \ceg~generates the C-Abstraction code for the original model and the quantized candidate and then checks their equivalence. Each check for this equivalence property is exported to a unique verification test case. Then, it triggers the execution of the verifier for each verification test case and awaits the verifier output. Here, \textit{Verifier module} is an instance of a formal verifier (i.e., a Bounded Model Checker (BMC), namely, ESBMC~\cite{Gadelha2018}). This step is done sequentially, meaning each verification is run once the last verification terminates. 

Once all verification test cases terminate, \ceg~collect and process all outputs and checks whether any counter-example has been found. If so, it updates the set of counter-examples~\ce, and triggers the GA module execution again, thus initiating a new iteration of \ceg. If no counter-example is found, \ceg~considers the verification successful and terminates the quantization process outputting the found solution.

We work with two functional versions of the \nn. The GA module works with a functional \nn~written in Python, while the verifier module works with a functional version of the \nn~written in C. The two models are equivalent since they share the same parameters; the python model loads the parameters to a framework built over Pytorch~\cite{NEURIPS2019_9015}. The C version loads the weights into a framework designed and developed in C to work correctly with the verifier idioms and annotations. We provide more details regarding the C implementations of the \ns~in Section~\ref{sec:c-abstractions}.


\section{Experimental Evaluation}
\label{sec:Experimental-Evaluation}

This section describes our experimental setup and benchmarks, defines our objectives, and presents the results. 

\subsection{Description of the Benchmarks}
\label{sec:benchmarks}

We evaluate our methodology on a set of feedforward \nn~classification models extracted from the literature~\cite{Eleftheriadis2022OnNN,Julian2016,LeCun2005TheMD}. We chose these specific ones based on their popularity in previous \nn~robustness and equivalence verification studies \cite{Teuber2021Equiv,Eleftheriadis2022OnNN}. Additionally, we include a few other \nn~models to cover a broader range of \nn~architectures (e.g., \nn~size, number of neurons).

\subsubsection{ACAS Xu}
The airborne collision avoidance system for unmanned aircraft ACAS Xu dataset~\cite{Julian2016} is derived from 8 specifications (features boundaries and expected outputs). ACAS Xu features are sensor data indicating the speed, present course of the aircraft, and the position and speed of any nearby intruder aircraft. An ACAS Xu \nn~ is expected to give appropriate navigation advisories for a given input sensor data. The expected outputs indicate that either the aircraft is clear-of-conflict, or it should take soft or hard turns to avoid the collision. We evaluated \ceg~on 5 pre-trained \ns, each containing 8 layers and 300 ReLU nodes each. The pre-trained \ns~were obtained from the VNN-COMP2021~\cite{bak2021second} benchmarks\footnotemark{}

\subsubsection{MNIST}
MNIST is a popular dataset~\cite{LeCun2005TheMD} for image classification. The dataset contains 70,000 gray-scale images with uniform size of 28x28 pixels, where the original pixel values from the integer range $[0,255]$ are rescaled to the floating-point range $[0,1]$. We evaluated \ceg~on two \ns~with 2 layers, one with 10 ReLU nodes each and another with 25 and 10 ReLU nodes. The \ns~followed the architecture of models described by the work of Eleftheriadis et al.~\cite{Eleftheriadis2022OnNN}.
 
 \footnotetext{The pre-trained weight for the ACAS Xu benchmarks can be found in the following repository: \url{https://github.com/stanleybak/vnncomp2021}}

\subsubsection{Seeds}
The Seeds dataset~\cite{Charytanowicz} consists of $210$ samples of wheat grain belonging to three different species, namely Kama, Rosa and Canadian. The input features are seven measurements of the wheat kernel geometry scaled between [0,1]. We evaluated \ceg~on 2 \ns, containing 1 layer, one containing 15 ReLU nodes, and the other containing 2 ReLU nodes. Both \ns~were trained for the \ceg~evaluation.

\subsubsection{Iris}

The Iris flower dataset~\cite{IrisDataset} consists of $50$ samples from three species of Iris flower (\textit{Iris setosa, Iris virginica and Iris versicolor}). The dataset is a popular benchmark in machine learning for classification, and the data is composed of records of real value measurements of the width and length of sepals and petals of the flowers. The data was scaled to [0,1]. We evaluated \ceg~on 2 \ns, one of them containing 2 layers with 20 ReLU nodes and the other having only one layer with 3 ReLU nodes. Both \ns~were trained for the \ceg~evaluation.

\subsection{Setup}

\subsubsection{Genetic Algorithm.}

As explained in Section~\ref{sec:robust-quantization}, we quantize the \ns~with a NSGA-II Genetic Algorithm module. We set the upper and lower bounds for the allowed bit widths to 2 and 52 in all experiments. The lower bound was chosen because 2 is the first valid integer that does not break our quantization formulas. The upper bound was chosen to match the significand of the double-precision float format IEEE 754-1985~\cite{8766229}. The upper bound value could be higher depending on the precision of weights parameters of the \nn, as the scaling factor could lead the quantization to large integer values. However, as we wanted the framework to work on every \nn~in our experimentation setup without further steps, we restricted the clipping range to a comfortable number to avoid integer overflow.  

Furthermore, we allow the GA to run for $110$ generations for each layer in the \nn. This number of generations was defined after extensive preliminary tests, which confirmed that GA could reach the optimal solution in most cases (see Table~\ref{tab:brute-force-bits-finding} in Appendix~\ref{sec:appendices-ga}). Lastly, we randomly select the initial set of counter-examples \mc{H} from the benchmark set of each case study. The samples in \mc{H} do not necessarily have to be \textit{counter-examples}, and any valid concrete input can be specified. Our choice is justified by the practical aspect of using samples from the benchmark set.

\subsubsection{Equivalence Properties.}

One input sample was selected for each output class and used to define the equivalence properties. Due to the high dimensional number of the features in the MNIST study case, we proposed a different approach when specifying the equivalence properties for the equivalence verification. We considered three different approaches:
1) one in which we considered all features in the input domain;
2) another one in which we considered only a subset of $10$ out of the $784$ features in the input domain;
3) a last one in which we considered only a subset of $4$ out of the $784$ features in the input domain.
The subset of features in cases $2$ and $3$ was randomly selected.


\subsubsection{Availability of Data and Tools.} Our experiments are based on a set of publicly available benchmarks. All tools, benchmarks, and results of our evaluation are available on a supplementary web page \url{https://zenodo.org/record/6791964}.

\subsection{Objectives}
\label{sec:objectives}

Considering the benchmarks given in Section~\ref{sec:benchmarks}, our evaluation has the following two experimental goals:

\begin{tcolorbox}
\begin{itemize}
    \item[EG1] (\textbf{robustness}) Show that the \ceg~framework can generate robust quantized \ns.
    \item[EG2] (\textbf{accuracy}) Show that the quantized \ns~do not have a significant drop in accuracy compared to other quantization techniques.
\end{itemize}
\end{tcolorbox}

\subsection{Results} 
\label{sec:results}

In our first set of experiments, we want to achieve our first experimental goal \textbf{EG1}. We want to show that our technique \ceg~can successfully generate quantized \ns~that are verifiably equivalent to the original \ns. As a secondary goal, we want to perform an empirical scalability study to help us evaluate the computational demands for quantizing and verifying the equivalence of \ns~models. Our findings are summarized in Table~\ref{tab:rq-cegion-summary}.

\begin{table}[ht!]
\centering
\caption{\label{tab:rq-cegion-summary} Summary of the \ceg~executions, including the models, number of features, the number of bits per layer, and the status.}
\begin{tabular}{lcccccc}
\toprule
 \textbf{Model}                       & \textbf{Features}  & \textbf{Equivalence Properties}   & \textbf{Iterations} & \textbf{Bits}                     & \textbf{Status}\\ \midrule
 iris\_3                     & 4         & 3                        & 1 & 4, 3                     & \emph{completed} \\
 seeds\_2                    & 7         & 3                        & 1 & 4, 3                     & \emph{completed} \\
 seeds\_15                   & 7         & 3                        & 1 & 4, 2                     & \emph{completed} \\ 
 acasxu\_1                   & 5         & 6                        & 1 & 6, 8, 7, 7, 9, 7, 6      & \emph{completed} \\ 
 acasxu\_2                   & 5         & 7                        & 1 & 10, 9, 9, 9, 7, 7, 10    & \emph{completed} \\ 
 acasxu\_3                   & 5         & 7                        & 1 & 5, 9, 10, 7, 8, 8, 5     & \emph{completed} \\ 
 acasxu\_4                   & 5         & 7                        & 1 & 8, 9, 14, 9, 10, 10, 7   & \emph{completed} \\ 
 acasxu\_5                   & 5         & 7                        & 1 & 6, 12, 8, 8, 10, 10, 10  & \emph{completed} \\
 \multirow{3}{*}{mnist\_10}  & 5         & 10                       & 1 & 4, 3                     & \emph{completed} \\
                             & 10        & 10                       & 1 & 4, 3                     & \emph{completed} \\
                             & 784       & 10                       & 0 & 4, 3                     & \emph{timeout} \\
\multirow{3}{*}{mnist\_25}   & 5         & 10                       & 1 & 3, 3                     & \emph{completed} \\
                             & 10        & 10                       & 1 & 3, 3                     & \emph{completed} \\
                             & 784       & 10                       & 0 & 3, 3                     & \emph{timeout} \\
\bottomrule
\end{tabular}
\end{table}

All the \ceg~runs that were completed successfully took only 1 iteration to find a solution. However, we observed that four of the \ceg~attempts to find a solution for MNIST models resulted in a timeout. We attribute this observation to a mix of factors. First is the high number of features in the MNIST problem. Second, the network's overall architecture requires many arithmetic operations to compute the model's output. Finally, we also observed that it took only a few minutes for \ceg to find a solution to the Iris, Seeds, and Acas Xu benchmarks. In contrast, on MNIST, it took hours to either find a solution or fail with a timeout.

\begin{tcolorbox}
These results answer our \textbf{EG1}: overall, these experiments show that CEG4N can successfully produce robust quantized models. Although, one should notice that for larger \ns~models, scalability should be a point of concern due to our verifier stage.
\end{tcolorbox}

In our second set of experiments, we want to achieve our second experimental goal \textbf{EG2}. We primarily want to understand the impact of the quantization performed by \ceg~on the accuracy of the \ns~ compared to other quantization techniques. Due to our research's novelty, no existing techniques lend themselves to a fair comparison. For this reason, we take a recent post-training quantization technique called GPFQ~\cite{zhang2022post} and modify it to our needs. GPFQ~\cite{zhang2022post} is a greedy path-following quantization technique that also produces quantized models with floating/double-precision values. It works by iterating over each layer of the \nn~and quantizing each neuron sequentially. More specifically, a greedy algorithm minimizes the error between the original neural output and the quantized neuron.

Table~\ref{table:compressed-accuracy} summarizes the accuracy of the models quantized using \ceg~and GPFQ. Note that we do not report the accuracy of the Acas Xu models because the original training and test datasets are not public.

\begin{table}[ht!]
\centering
\caption{Comparison of Top-1 accuracy for \ns~quantized using \ceg~and GPFQ\label{table:compressed-accuracy}}
\begin{tabular}{lcccc}
 \toprule
 Model                          & Method   & Ref Acc (\%)           & Quant Acc (\%) & Acc Drop (\%)\\
 \midrule
 \multirow{2}{*}{iris\_3}       & \ceg~& \multirow{2}{*}{93.33} & 83.33         & 10.0          \\
                                & GPFQ     &                        & 23.33         & 70.0          \\
 \multirow{2}{*}{seeds\_2}      & \ceg~& \multirow{2}{*}{88.09} & 85.71         & 2.38          \\
                                & GPFQ     &                        & 64.28         & 23.81         \\
 \multirow{2}{*}{seeds\_15}     & \ceg~& \multirow{2}{*}{90.04} & 85.71         & 4.33          \\
                                & GPFQ     &                        & 40.47         & 49.57         \\
 \multirow{2}{*}{mnist\_10}     & \ceg~& \multirow{2}{*}{91.98} & 86.7          & 5.28          \\
                                & GPFQ     &                        & 91.29         & 0.69          \\
  \multirow{2}{*}{mnist\_25}    & \ceg~& \multirow{2}{*}{93.68} & 92.57         & 1.11          \\
                                & GPFQ     &                        & 92.59         & 1.09          \\
 \bottomrule
\end{tabular}
\end{table}

Our findings show that the highest drops in accuracy happen on the Iris benchmark (10\% for \ceg~and 70\% drop for GPFQ). In contrast, the lowest drops in accuracy happen on mnist\_25 for \ceg~and on mnist\_10 for GPFQ. Overall, the accuracy of models quantized with \ceg~are better on the Iris and Seeds benchmarks, while the accuracy of models quantized with GPFQ are better on the mnist benchmarks, but only by a small margin. Our understanding is that GPFQ shows high drops in accuracy for smaller \ns~because the number of neurons in each layer is small. As GPFQ focuses on each neuron individually, it may not be able to find a good global quantization. 

\begin{tcolorbox}
These results answer our \textbf{EG2}: overall, these experiments show that \ceg~can successfully produce quantized models with superior or similar accuracy to other state-of-the-art techniques.
\end{tcolorbox}

\subsection{Limitations}
\label{sec:ThreatstoValidity}
Although we showed in our evaluation that the \ceg~framework can generate a quantized neural network while keeping the equivalence between the original \nn~and the quantized \nn, we note that the architecture of the \nn~used in the evaluation does not fully reflect state-of-the-art \nn~architectures. The \ns~used in our evaluation have few layers and only hundreds of ReLU nodes, while state-of-the-art \ns~may have hundreds of layers and thousands of ReLU nodes. The main bottleneck is state-of-the-art verification algorithms, which currently do not scale to large neural networks. As it is, our technique could only quantized 80\% of the \nn~in our experimental evaluation.

In addition, the field of research on \nn~equivalence is relatively new and there is no well-established set of benchmarks that works in this field could benefit from~\cite{Eleftheriadis2022OnNN}. Furthermore, our work is the first to propose a framework that mixes \nn~quantization and \nn~equivalence verification. There is no comparable methodology in the literature we could compare our approach with.


\section{Conclusion}
\label{sec: Conclusion}

We presented a new method for \nn~quantization, called \ceg, a post-training \nn~quantization technique that provides formal guarantees of \nn~equivalence. This approach leverages a counter-example guided optimization technique, where an optimization-based quantizer produces quantized model candidates. A state-of-the-art C verifier then checks these candidates to prove the equivalence of the quantized candidates and the original models or refute that equivalence by providing a counter-example. This counter-example is then passed back to the quantized to guide it to search for a feasible candidate. 

We evaluate the \ceg~method on four benchmarks, including large models (ACAS Xu and MNIST) and smaller models (Iris and Seeds). We successfully demonstrate the application of the \ceg~for \nn~quantization, where it could successfully quantize the networks while producing models with up to 72\% better accuracy than state-of-the-art techniques. However, \ceg~can only handle a restricted set of \ns~models, and further work needs to scale the \ceg~applicability on a broader set of \ns~models (e.g., \ns~models with a more significant number of layers and neurons and higher numbers of input features). 

For future work, we could explore other quantization techniques, which are not limited to search-based quantization and other promising equivalence verification techniques using a MILP approach~\cite{Teuber2021Equiv} or an SMT-based approach~\cite{Eleftheriadis2022OnNN}. Combining different quantization and equivalence verification techniques can enable \ceg~to achieve better scalability and quantization rates. Another interesting future work relates to the possibility of mixing quantization approaches that generate quantized models, which operate entirely on integer arithmetic; this can potentially improve the verification step scalability of the \ceg. 

\section*{Acknowledgment}
The work is partially funded by EPSRC grant EP/T026995/1 entitled ``EnnCore: End-to-End Conceptual Guarding of Neural Architectures'' under \textit{Security for all in an AI-enabled society}.

\bibliographystyle{splncs04}
\bibliography{main.bib}

\appendix

\section{Appendices}
\label{sec:appendices}



\subsection{Implementation of \ns~in Python.}
The \ns~were built and trained using the Pytorch library~\cite{NEURIPS2019_9015}. Weights of the trained models were then exported to the ONNX~\cite{bai2019} format, which can be interpreted by Pytorch and used to run predictions without any compromise in the \ns~performance. 

\subsection{Implementation of \ns~abstract models in C.}
\label{sec:c-abstractions}

In the present work, we use the C language to implement the abstract representation of the \ns. It allows us to explicitly model the \nn~operations in their original and quantized forms and apply existing software verification tools (e.g., ESBMC~\cite{GadelhaMC21}). The operational C-abstraction models perform double-precision arithmetic. Although, we must notice that the original and quantized only diverge on the precision of the weight and bias vectors that are embedded in the abstractions code.


\subsection{Encoding of Equivalence Properties}

Suppose, a \nn~ $F$, for which $x\in \mc{H}$ is a safe input and $y \in \mc{G}$ is the expected output of $f$ the input. We now show how one can specify the equivalence properties. For this example, consider that the function $f$ can produce the outputs of $F$ in floating-point arithmetic, while $fq$ produces the outputs of $F$ in fixed-point arithmetic (\ie~quantization). First, the concrete \nn~ input $x$ is replaced by a non-deterministic one, which is achieved using the command \textbf{nondet\_float} from the ESBMC.

\begin{lstlisting}[language=C,tabsize=1,firstline=8, lastline=12, caption=Definition of concrete and symbolic input domain in \emph{EBMC}.]
	float x0 = -1.0;
	float x1 = 1.0;
	float s0 = nondet_float();
	float s1 = nondet_float();
	
\end{lstlisting}


\begin{lstlisting}[language=C,tabsize=1,firstline=13, lastline=16, caption=Definition of input constraints in \emph{EBMC}.]
	const float EPS = 0.5;
	__ESBMC_assume(x0 - EPS <= s0 && s0 <= x0 + EPS);
	__ESBMC_assume(x1 - EPS <= s1 && s1 <= x1 + EPS);
	
\end{lstlisting}

\begin{lstlisting}[language=C,tabsize=1,firstline=17, lastline=18, caption=Definition of output constraints in \emph{EBMC}.]
	__ESBMC_assert(f(s0, s1) == fq(s0, s1));
	
\end{lstlisting}

\subsection{Genetic Algorithm Parameters Definition}
\label{sec:appendices-ga}

In Table~\ref{tab:brute-force-bits-finding}, we report a summary of experiments conducted to tune the parameters of the Genetic Algorithm, more precisely, the number of generations. For example, a \nn~with $2$ layers would require a brute force algorithm to search for $52^2$ combinations of bits widths for the quantization. Similarly, a \nn~with $7$ layers would require a brute force algorithm to search for $52^7$ combinations of bits widths. We conducted a set of experiments where we ran the GA one hundred times with a different number of generations options ranging from $50$ to $1000$. In addition, we fixed the population size to $5$. From our findings, the GA needs about $100$ to $110$ generations per layer to find the optimal bit width solution for each run.

\begin{table}[h!]
    \centering
    \caption{Summary of experiments for tuning Genetic Algorithm Parameters.\label{tab:brute-force-bits-finding}}
    \begin{tabular}{cccc}
        \toprule
        \textbf{Number of Layers} & \textbf{Generations} & \textbf{Population} & \textbf{Percentage of optimal solutions}\\
        \midrule
        7 & 800 & 5 & 100\\
        7 & 750 & 5 & 100\\
        7 & 700 & 5 & 98\\
        7 & 50 & 5 & 0\\
        
        2 & 250 & 5 & 100\\
        2 & 200 & 5 & 100\\
        2 & 150 & 5 & 96\\
        2 & 50 & 5 & 30\\
        \bottomrule
    \end{tabular}
\end{table}
\end{document}